\documentclass[11pt,a4paper]{article}
\pdfoutput=1
\usepackage{times}
\usepackage{latexsym}
\usepackage{microtype}
\usepackage{booktabs}
\usepackage{multirow}
\usepackage{graphicx}
\usepackage{listings}
\usepackage{amsmath}

\usepackage{stfloats}
\usepackage{verbatim}
\usepackage{subcaption}

\usepackage{xurl}
\usepackage[hyperref]{emnlp2020}

\aclfinalcopy

\title{Literature Retrieval for Precision Medicine \\
with Neural Matching and Faceted Summarization}

\author{Jiho Noh \\
  Department of Computer Science \\
  University of Kentucky\\
  Kentucky, USA \\
  \texttt{jiho.noh@uky.edu} \\ \And
  Ramakanth Kavuluru \\
  Division of Biomedical Informatics\\
  University of Kentucky\\
  Kentucky, USA \\
  \texttt{ramakanth.kavuluru@uky.edu}\\ }

\date{\today}

\begin{document}
\maketitle

\begin{abstract}
Information retrieval (IR) for precision medicine (PM) often involves looking for multiple pieces of evidence that characterize a patient case.
This typically includes at least the name of a condition and a genetic variation that applies to the patient. Other factors such as demographic attributes, comorbidities, and social determinants may also be pertinent. As such, the retrieval problem is often formulated as \textit{ad hoc} search but with multiple facets (e.g., disease, mutation) that may need to be incorporated. In this paper, we present a document reranking approach that combines neural query-document matching and text summarization toward such retrieval scenarios.
Our architecture builds on the basic BERT model with three specific components for reranking: (a). document-query matching (b). keyword extraction and (c). facet-conditioned abstractive summarization.
The outcomes of (b) and (c) are used to essentially transform a candidate document into a concise summary that can be compared with the query at hand to compute a relevance score. Component (a) directly generates a matching score of a candidate document for a query.
The full architecture benefits from the complementary potential of document-query matching and the novel document transformation approach based on summarization along PM facets. Evaluations using NIST's TREC-PM track datasets (2017--2019) show that our model achieves state-of-the-art performance. To foster reproducibility, our code is made available here: \url{https://github.com/bionlproc/text-summ-for-doc-retrieval}.
\end{abstract}

\section{Introduction}
The U.S.~NIH's precision medicine (PM) initiative~\cite{collins2015new} calls for designing treatment and preventative interventions considering genetic, clinical, social, behavioral, and environmental exposure variability among patients. The initiative rests on the widely understood finding that considering individual variability is critical in tailoring healthcare interventions to achieve substantial progress in reducing disease burden worldwide. Cancer was chosen as its near term focus with the eventual aim of expanding to other conditions.
As the biomedical research enterprise strives to fulfill the initiative's goals, computing needs are also on the rise in drug discovery, predictive modeling for disease onset and progression, and in building NLP tools to curate information from the evidence base being generated.

\subsection{TREC Precision Medicine Series}
\begin{table}[hbt]
  \centering
  \begin{tabular}{@{}ll@{}}
    \toprule
    Facet & Input\\
     \midrule
    Disease & Melanoma \\
    Genetic variation & BRAF (E586K) \\
    Demographics & 64-year-old female \\
    \midrule
    Disease & Gastric cancer \\
    Genetic variation & ERBB2 amplification\\
    Demographics & 64-year-old male\\
    \bottomrule
  \end{tabular}
  \caption{Example   cases from  2019 TREC-PM dataset~\label{tbl:trecpm_ex_topics}}
\end{table}

In a dovetailing move, the U.S.~NIST's  TREC (Text REtrieval Conference) has been running a PM track since 2017 with a focus on cancer~\cite{roberts2020overview}. The goal of the TREC-PM task is to identify the most relevant biomedical articles and clinical trials for an input patient case.
Each case is composed of   (1) a disease name,  (2) a gene name and genetic variation type, and (3) demographic information (sex and age). Table~\ref{tbl:trecpm_ex_topics} shows two example cases from the 2019 track. So the search is \textit{ad hoc} in the sense that we have a free text input in each facet but  the    facets themselves highlight the PM related attributes that ought to characterize the retrieved documents. We believe this style of faceted retrieval is going to be more common across medical IR tasks for many conditions as the PM initiative continues its mission.

\subsection{Vocabulary Mismatch and Neural IR}

The vocabulary mismatch problem is a prominent issue in medical IR given the large variation in the expression of medical concepts and events. For example, in the query \textit{``What is a potential side effect for Tymlos?''} the drug is referred by its brand name. Relevant scientific literature may contain the generic name \textit{Abaloparatide} more frequently. Traditional document search engines have clear limitations on resolving   mismatch issues.
The IR community has extensively explored methods to address the \textit{vocabulary mismatch} problem, including query expansion based on relevance feedback, query term re-weighting, or query reconstruction by optimizing the query syntax.

Several recent studies highlight exploiting neural network models for query refinement in document retrieval (DR) settings. \citet{nogueira2017task}  address  this issue by generating a transformed query from the initial query using a neural model.  They use reinforcement  learning (RL) to train it where an \textit{agent} (i.e., reformulator) learns to reformulate the initial query to maximize the expected return (i.e., retrieval performance) through \textit{actions} (i.e., generating a new query from the output probability distribution). In a different approach, \citet{narayan2018ranking}  use RL for sentence ranking for extractive summarization.

\subsection{Our Contributions}
\label{sec-contrib}
In this paper, building on the BERT architecture~\cite{devlin2019bert}, we focus on a different hybrid document scoring and reranking setup involving three components: (a).~a \textit{document relevance classification} model, which predicts (and inherently scores) whether a document is relevant to the given query (using a BERT multi-sentence setup); (b).~a \textit{keyword extraction} model which spots tokens in a document that are likely to be seen in PM related queries; and (c).~an \textit{abstractive document summarization} model that generates a pseudo-query given the document context and a facet type (e.g., genetic variation) via the BERT encoder-decoder setup.
The keywords (from (b)) and the pseudo-query (from (c)) are together compared with the original query to generate a score.
The scores from all the components are combined to rerank top $k$ (set to 500) documents returned with a basic Okapi BM25 retriever from a Solr index~\cite{grainger2014solr} of the corpora.

Our main innovation is in pivoting from the focus on queries by previous methods to emphasis on transforming candidate documents into pseudo-queries via summarization. Additionally, while generating the pseudo-query, we also let the   decoder output concept codes from biomedical terminologies that capture disease and gene names. We do this by embedding both words and concepts in a common semantic space before letting the decoder generate summaries that include concepts.
Our overall architecture was evaluated using the TREC-PM datasets (2017--2019) with the 2019 dataset used as the test set.
The results show an absolute $4\%$ improvement in P@10 compared to prior best approaches while obtaining a small $\approx 1\%$ gain in R-Prec. Qualitative analyses also highlight how the summarization is able to focus on document segments that are highly relevant to patient cases.

\section{Background}
The basic reranking architecture we begin with is  the
Bidirectional Encoder Representations from Transformers (BERT)~\cite{devlin2019bert} model. BERT is trained on a \textit{masked language modeling} objective on a large text corpus such as \textit{Wikipedia} and \textit{BooksCorpus}. As a sequence modeling method, it has achieved state-of-the-art results in a wide range of natural language understanding (NLU) tasks, including machine translation~\cite{conneau2019cross} and text summarization~\cite{liu2019text}.
 With an additional layer on top of a pretrained BERT model, we can fine-tune models for specific NLU tasks. In our study, we utilize this framework in all three components identified in Section~\ref{sec-contrib} by starting with a \texttt{bert-base-uncased} pretrained   HuggingFace model~\cite{wolf2019transformers}.

\subsection{Text Summarization}
We plan to leverage both extractive and abstractive candidate document summarization in our framework.
In terms of learning methodology, we view extractive summarization as a sentence (or token) classification problem. Previously proposed models include the RNN-based sequence model~\cite{nallapati2017summarunner}, the attention-based neural encoder-decoder model~\cite{cheng2016neural}, and the sequence model with a global learning objective (e.g., ROUGE) for ranking sentences optimized via RL~\cite{narayan2018ranking,paulus2017deep}. More recently,  graph convolutional neural networks (GCNs) have also been adapted to allow the incorporation of global information in text summarization tasks~\cite{sun2019divgraphpointer,prasad2019glocal}.
Abstractive summarization is typically cast as a sequence-to-sequence learning problem. The encoder of the framework reads a document and yields a sequence of continuous representations, and the decoder generates the target summary token-by-token~\cite{rush2015neural,nallapati2016abstractive}.
Both approaches have their own merits in generating comprehensive and novel summaries; hence most systems leverage these two different models in one framework~\cite{see2017get,liu2019text}. We use the extractive component to identify tokens in a candidate document that may be relevant from a PM perspective and use the abstractive component to identify potential terms that may not necessarily be in the document but nevertheless characterize it for PM purposes.

\subsection{Word and Entity Embeddings ~\label{sec:word_embeddings}}
Most of the neural text summarization models, as described in the previous section, adopt the encoder-decoder framework that is popular in machine translation. As such the vocabulary on the decoding side does not have to be the same as that on the encoding side. We exploit this to design a summarization trick for PM where the decoder outputs both regular English tokens and also entity codes from a standardized biomedical terminology that captures semantic concepts discussed in the document. This can be trained easily by converting the textual queries in the training examples to their corresponding entity codes. This trick is to enhance our ability to handle vocabulary mismatch in a different way (besides the abstractive framing).
 We created \textit{BioMedical Entity Tagged} (BMET) embeddings\footnote{\url{https://github.com/romanegloo/BMET-embeddings}} for this purpose. BMET embeddings are trained on biomedical literature abstracts that were annotated with entity codes in the  Medical Subject Headings (MeSH) terminology\footnote{\url{https://www.nlm.nih.gov/mesh/meshhome.html}}; codes are appended to the associated textual spans in the training examples. So regular tokens and the entity codes are thus embedded in the same semantic space via pretraining with the \textit{fastText} architecture~\cite{bojanowski2017enriching}. Besides regular English tokens, the vocabulary of BMET thus includes 29,351 MeSH codes and a subset of supplementary concepts.
 In the dictionary, MeSH codes are differentiated from the regular words by a unique prefix; for example, \textit{$\epsilon$mesh\_d000123} for MeSH code \textit{D000123}. With this, our summarization model can now translate a sequence of regular text tokens into a sequence of biomedical entity codes or vice versa. That is, we use MeSH as a new ``semantic'' facet besides those already provided by TREC-PM organizers. The expected output for the MeSH facet is the set of codes that capture entities in the disease and gene variation facets.

\section{Models and Reranking}

\begin{figure*}[t]
  \centering
  \includegraphics[scale=0.42]{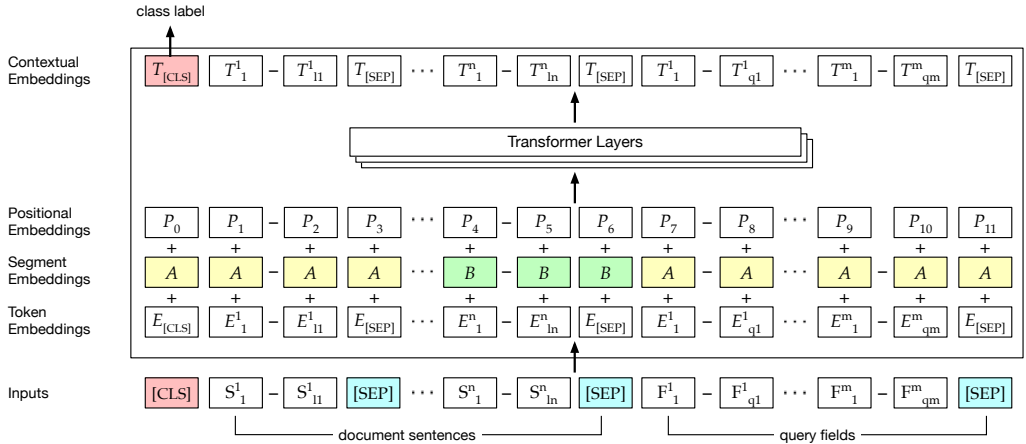}
  \caption{BERT architecture for document relevance matching task \texttt{REL}\label{fig:encoder}}
\end{figure*}

In this effort, toward document reranking, we aim to measure the relevance match between a document and a faceted PM query. Each training instance is a 3-tuple  ($d$, $q$, $y^d_q$) where $q$ is a query, $d$ is a candidate  document, and $y^d_q$ is a Boolean human adjudicated outcome: whether $d$ is relevant to $q$. As mentioned in Section~\ref{sec-contrib}, first, we fine-tune BERT for a  query-document relevance matching task modeled as a classification goal to predict $y^d_q$ (\texttt{REL}).   Next, we fine-tune BERT for token-level relevance classification, different from \texttt{REL}, where a token in $d$ is deemed relevant during training if it occurs as part of $q$. We name this model \texttt{EXT} for keyword extraction. Lastly, we train a BERT model in  the seq2seq  setting where the encoder is initialized with a pretrained \texttt{EXT} model. The encoder  reads in  $d$, and the decoder attends to the contextualized representations of $d$ to generate a facet-specific pseudo-query sentence $q_d$, which is then compared with the original query $q$. We conceptualize this process as  text summarization from a document to query sentences\footnote{We note queries here are not grammatically well-formed sentences but are essentially sequences generated by the summarization model.} and refer to it as \texttt{ABS}. All three models are used together to rerank a candidate $d$ at test time for a specific input query.

\subsection{Document Relevance Matching (\texttt{REL})}
Neural text matching has been recently carried out through siamese style networks~\cite{mueller2016siamese}, which also have been adapted to biomedicine~\cite{noh2018document}. Our approach adapts the   BERT architecture for the matching task in the multi-sentence setting as shown in Figure~\ref{fig:encoder}.
We use BERT's tokenizer on its textual inputs, and the tokens are mapped to token embeddings. REL takes the concatenated sequence of a document and faceted query sentences. The functional symbols defined in the BERT tokenizer (e.g., \texttt{[CLS]}) are added to the input sequence. Each input sequence starts with a \texttt{[CLS]} token. Each sentence of the document ends with the \texttt{[SEP]} token with the last segment of the input sequence being the set of faceted query sentences, which end with another \texttt{[SEP]} token. In the encoding process, the first \texttt{[CLS]} token collects features for determining document relevance to the query.
BERT uses segment embeddings to distinguish two sentences. We, however, use the them to distinguish multiple sentences within a document. For each sentence, we assign a segment embedding either $A$ or $B$ alternatively. The positional embeddings encode the sequential nature of the inputs. The token embeddings along with the segment and positional embeddings pass through the transformer layers. Finally, we use the $[0,1]$ output logit from the \texttt{[CLS]} token ($T_{[CLS]}$) as the matching score for the input document and query.
We note that we don't demarcate any boundaries within different facets of the query.

\vspace{-2mm}
\subsection{Keyword Extraction (\texttt{EXT})}

\texttt{EXT} model has an additional token classification layer on top of the pretrained BERT. The output of a token is the logit that indicates the log of odds of the token's occurrence in the query. With   TREC-PM datasets, we expect to see the logits fire for words related to different facets with an optimized \texttt{EXT} at test time. Unlike the \texttt{REL} model, the input to \texttt{EXT} is a sequence of words in a document without any  \texttt{[SEP]} delimiters. However, the model still learns the boundaries of the sentence via segment inputs. This component essentially generates a brief extractive summary of a candidate document. Furthermore,  contextualized embeddings  from \texttt{EXT} are used in the decoder of \texttt{ABS} to generate faceted abstractive document summaries .

\begin{figure*}[t]
  \centering
  \includegraphics[scale=0.4]{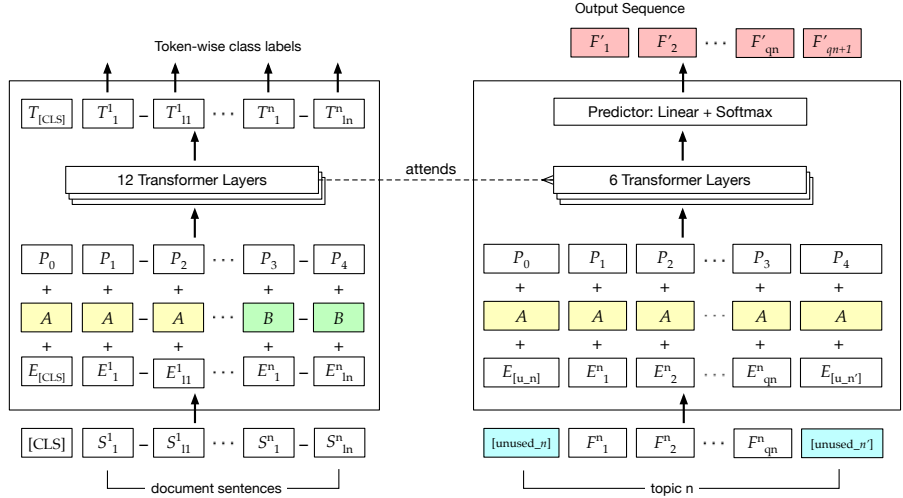}
  \caption{Architecture of the abstractive document summarization (\texttt{ABS}) model. The encoder (left component) is initialized with a pretrained \texttt{EXT} model. The class labels of the encoder are used for identifying keywords of the document, and the output sequences generated from the decoder (right component) are used to build a pseudo-query, which is later used in computing similarity scores for the user provided query.\label{fig:abs}}
\end{figure*}

\subsection{Abstractive Document Summarization (\texttt{ABS})}

\texttt{ABS} employs a standard seq2seq attention model, similar to that  by \citet{nallapati2016abstractive}, as shown in  Figure~\ref{fig:abs}. We initialize the parameters of the encoder with a pretrained \texttt{EXT} model. The decoder is a 6-layer transformer in which the self-attention layers attend to only the earlier positions in the output sequence as is typical in  auto-regressive language models. In each training phase step, the decoder takes each previous token from the reference query sentence; in the generation process, the decoder uses the token predicted one step earlier.

\begin{table}[hb]
  \centering
  \small
  \begin{tabular}{@{}lc@{}}
    \toprule
    Facets & (bos)/(eos)\\
    \midrule
    Disease name & \texttt{[unused\_0]/[unused\_100]}\\
    Genetic variations & \texttt{[unused\_1]/[unused\_101]}\\
    Demographic info.& \texttt{[unused\_2]/[unused\_102]}\\
    MeSH terms & \texttt{[unused\_3]/[unused\_103]}\\
    Document keywords & \texttt{[unused\_4]/[unused\_104]}\\
    \bottomrule
  \end{tabular}
  \caption{Signals for different facets of the patient cases~\label{tbl:field_signals}}
\end{table}

We differentiate facets by the special pairs of tokens assigned to each topic. In a typical generation process,   special tokens such as \texttt{[bos]} (begin) and \texttt{[eos]} (end) are used to indicate sequence boundaries. In this model, we use some special tokens in the BERT vocabulary with prefix `\texttt{unused\_}'.  Specifically, \texttt{[unused\_$i$]} and \texttt{[unused\_$(100+i)$]}  are used as \texttt{bos} and \texttt{eos} tokens respectively for different facets. These facet signals are the latent variables for which  \texttt{ABS}   is optimized. Through them, \texttt{ABS} learns not only the thematic aspects of the queries but also the meta attributes such as length. The special tokens for   facets are listed in Table~\ref{tbl:field_signals} (the last row indicates a new auxiliary facet  we introduce in Section~\ref{sec:data}).

Each faceted query   is enclosed by its assigned \texttt{bos}/\texttt{eos} pair, and the decoder of \texttt{ABS} learns $p_\theta (x_i | x_{< i}, x_0)$ , where $x_0$ is the facet signal. As  in the encoder and the original transformer architecture~\cite{vaswani2017attention}, we add the sinusoidal positional embedding $P_t$ and the segment vector $A$ (or $B$) to the token embedding $E_t$. Note that the dimension of the token embeddings used in the encoder (BERT embeddings) is different from that of the decoder (our custom BMET embeddings), which causes a discrepancy in computing   context-attentions of the target text across the source document. Hence, we add an additional linear layer to project the constructed decoder embeddings ($E^n_j+A+P_i$ in the right hand portion of Figure~\ref{fig:abs}) into the same space of embeddings of the encoder.
These projected   embeddings   are fed to the  decoder's transformer layers. Each transformer layer applies multi-head attention for computing the self- and context-attentions. The attention function reads the input masks to preclude attending to future tokens of the input and any padded tokens (i.e., \texttt{[PAD]}) of the source text. Both attention functions apply a residual connection~\cite{he2016deep}. Lastly, each transformer layer ends with a position-wise feedforward network. Final scores for each token are computed from the linear layer on top of the  transformer layers. In training, these scores are consumed by a cross-entropy loss function. In generation process, the softmax function is applied over the vocabulary yielding a probability distribution for sampling the next token.



Finally to generate the pseudo-query, we use  \textit{beam search} to find the most probable sentence among predicted candidates. The scores are penalized by two measures  proposed by \citet[Equation 14]{wu2016google}: (1). The length penalty
$ lp(Y) = (5 + |Y|)^\alpha/(5 + 1)^\alpha$,
where $|Y|$ is the current target length and  $0 < \alpha < 1$ is the length normalization coefficient. (2). The coverage penalty
$$ cp(X, Y) = \beta \sum_{i=1}^{|X|} \log(\mbox{min}(\sum_{j=1}^{|Y|} p_{i,j}, 1.0)),$$
where $p_{i,j}$ is the attention score of the $j$-th target word $y_j$ on the $i$-th source word $x_i$, $|X|$ is the source length, and $0 < \beta < 1$ is the coverage normalization coefficient. Intuitively, these functions avoid favoring shorter predictions and yielding duplicate terms. We tune the parameters of the penalty functions ($\alpha = \beta = 0.4$), with grid-search on the validation set for TREC-PM.

\subsection{Reranking with \texttt{REL}, \texttt{EXT}, and \texttt{ABS}}
\label{sec-rerank}
The main purpose of the models designed in the previous subsections is to come up with a combined measure for reranking.
For a query $q$, let $d_1, \ldots, d_r$, be the set of top $r$ (set to 500) candidate documents returned by the Solr BM25 eDisMax query.
It is straightforward to impose an order on $d_j$ through \texttt{REL} via the output probability estimates of relevance.
Given $q$, for each $d_j$ we generate the pseudo-query (summary)  $q_{d_j}$ by
concatenating all  distinct words in the generated pseudo-query sentences by \texttt{ABS} along with the words selected  by \texttt{EXT}.
Repeating words and special tokens are removed. Although faceted summaries are generated through ABS, in the end
$q_{d_j}$ is essentially the set of all unique terms from \texttt{ABS} and \texttt{EXT}.
Each $d_j$ is now scored by comparing $q$ and $q_{d_j}$ via two similarity metrics: The ROUGE-1 recall score, $s_{ROUGE}$~\cite{lin-2004-rouge}, and a cosine similarity based score computed as
\[ s_{\cos} (q, q_{d_j}) = \frac{1}{|q|} \sum_{y\in q} \max_{x\in q_{d_j}}(\cos(e_y, e_x)),\]
where $e_i$ denote  vector representations from BMET embeddings (Section~\ref{sec:word_embeddings}).

Overall, we compute four different scores (and hence rankings) of a document: (1) the retrieval score returned by Solr, (2) the document relevance score by \texttt{REL}, (3) pseudo-query based ROUGE score, and (4) pseudo-query similarity score $s_{\cos}$. In the end we merge the rankings with \textit{reciprocal rank fusion}~\cite{cormack2009reciprocal} to obtain the final ranked list of documents. The results are compared against the state-of-the-art models from the 2019 TREC-PM task.
\section{Experimental Setup}

\subsection{Data}\label{sec:data}
Across 2017--2019 TREC-PM tasks, we have a total of 120 patient cases and 63,387 qrels (document relevance judgments) as shown in  Table~\ref{tbl:trecpm_datasets}.

\begin{table}[htb]
  \centering
      \renewcommand{\arraystretch}{1.25}
  \begin{tabular}{@{}crrr@{}}
    \toprule
    Year & Queries & Documents (rel. / irrel.) \\
    \midrule
    2017 & 30 & 3,875 / 18,767 \\
    2018 & 50 & 5,588 / 16,841 \\
    2019 & 40 & 5,544 / 12,772 \\
    \bottomrule
  \end{tabular}
  \caption{Number of queries and pooled relevance judgments in the 2017--19 TREC-PM tracks~\label{tbl:trecpm_datasets}}
\end{table}

We create two new auxiliary facets, \textit{MeSH terms} and \textit{Keywords}, derived from  any training query and document pair. We already covered the MeSH facet in Section~\ref{sec:word_embeddings}. \textit{Keywords} are those assigned by authors to a biomedical article to capture its themes  and are downloadable from NIH's NCBI website.  If  no keywords were assigned to an article, then we use the set of preferred names of MeSH terms (assigned to the articles by trained NIH coders) for that example. The following list shows associated facets for a sample training instance:
\begin{itemize}
  \setlength\itemsep{-.2em}
  \item \textbf{Disease}: prostate cancer
  \item \textbf{Genetic variations}: ATM deletion
  \item \textbf{Demographics}: 50-year-old male
  \item \textbf{MeSH terms}: D011471, D064007
  \item \textbf{Keywords}: Aged, Ataxia Telangiectasia mutated Proteins, Prostate Neoplasms/genetics
\end{itemize}

Each model consumes data differently, as shown in Table~\ref{tbl:data_inputs}. \texttt{REL} takes a document along with the given query  as the source input and predicts document-level relevance. We consider a document with the human judgment score either 1 (partially relevant) or 2 (totally relevant) as relevant for this study. Note that we do not include MeSH terms in the query sentences for \texttt{REL}. \texttt{EXT} reads in a document as the source input and predicts token-level relevances. During training, a relevant token is one that occurs in the given patient case. A pseudo-query is the output for \texttt{ABS} taking in a document and a facet type.

\begin{table}[htb]
  \centering
    \renewcommand{\arraystretch}{1.25}
  \begin{tabular}{@{}ccc@{}}
    \toprule
    Model &  Source & Target \\
    \midrule
    REL & doc$+${query\_sentences} & doc relevance \\
    EXT & doc & token relevances \\
    ABS & doc$+${facet\_signal} & a pseudo-query  \\
    \bottomrule
  \end{tabular}
  \caption{Data inputs and outputs for each model.~\label{tbl:data_inputs}}
\end{table}

\vspace{-2mm}
\subsection{Implementation Details}

For all three models, we begin with the  pretrained \texttt{bert-base-uncased} HuggingFace model~\cite{wolf2019transformers}  to encode source texts. We use   BERT's \textit{WordPiece}~\cite{schuster2012japanese} tokenizer for the source documents.

\texttt{REL} and \texttt{EXT}  are trained for 30,000 steps with batch size of 12. The maximum number of tokens for source texts is limited to 384. As the loss function of these two models, we use \textit{weighted} binary cross entropy. That is, given high imbalance with many more irrelevant instances than positive ones, we put different weights on the classes in computing the loss according to the target distributions (proportions of negative examples are 87\% for \texttt{REL} and 93\% for \texttt{EXT}). The loss is
\[
l (x, y; \theta) = - w_y [ y   \log p(x) + ( 1 - y )   \log (1 - p(x)) ] ,
\]
where $w_0=13/87=0.15, w_1=1$ for \texttt{REL} and $w_0=7/93=0.075, w_1=1$ for \texttt{EXT}. Adam optimizer with  parameters $\beta_1=0.9$ and $\beta_2=0.999$, starting learning rate $lr=1e^{-5}$, and fixed weight decay of 0.0 was used. The learning rate is reduced when a metric has stopped improving by using the \textit{ReduceLROnPlateau} scheduler in \textit{PyTorch}.

For the decoder of \texttt{ABS}, multi-head attention module from OpenNMT~\cite{klein2017opennmt} was used. To tokenize  target texts, we use the  NLTK word tokenizer (\url{https://www.nltk.org/api/nltk.tokenize.html}) unlike the one used in the encoder; this is because we use customized word embeddings, the \textit{BMET} embeddings (Section~\ref{sec:word_embeddings}),  trained with a domain-specific corpus and vocabulary. The vocabulary size is 120,000 which includes the 29,351 MeSH codes. We use six transformer layers in the decoder. Model dimension is 768 and the feed-forward layer size is 2048. We use different initial learning rates for the encoder and decoder, since the encoder is initialized with a pretrained \texttt{EXT} model: $1e^{-5}$ (encoder) and $1e^{-3}$ (decoder).  Negative log-likelihood  is the loss function for \texttt{ABS} on the ground-truth faceted query sentences.
For beam search in \texttt{ABS}, \texttt{beam\_size} is set to 4. At test time, we select top two best predictions and merge them into one query sentence. The max length of target sentence is limited to 50 and a sequence is incrementally generated until \texttt{ABS} outputs the corresponding \texttt{eos} token for each facet.
All parameter choices were made based on best practices from prior efforts and experiments to optimize P@10 on validation subsets.

\section{Evaluations and Results}\label{sec:results}
We conducted both quantitative and qualitative evaluations with example outcomes. The final evaluation was done
on the 2019 TREC-PM dataset while all   hyperparameter tuning was done using a training and validation dataset split
of a shuffled combined set of instances from 2017 and 2018 tracks (20\% validation and the rest for training).

\subsection{Quantitative Evaluations}

We first discuss the performances of the constituent  \texttt{REL}  and  \texttt{EXT} models that were evaluated using train and validation splits from 2017--2018 years.
Table~\ref{tbl:cls_rst} shows their performance where \texttt{REL} can recover $\approx 92\%$ of the relevant documents and \texttt{EXT} can identify $\approx 88\%$ of the tokens that occur in patient case information, both at precisions over 90\%.  We find that learning a model for identifying document/token-level relevance is relatively straightforward even with the imbalance.

\begin{table}[hbt]
  \centering
  \small
    \renewcommand{\arraystretch}{1.3}

  \setlength{\tabcolsep}{3pt}
  \begin{tabular}{@{}ccccccc@{}}
    \toprule
    & \multicolumn{3}{c}{\texttt{REL}} & \multicolumn{3}{c}{\texttt{EXT}} \\
    \cmidrule(lr){2-4} \cmidrule(l){5-7}
    & P & R & F1 & P & R & F1 \\
    \midrule
    Train & 0.9814 & 0.9384 & 0.9594 & 0.9624 & 0.8877 & 0.9236\\
    Valid & 0.9266 & 0.9147 & 0.9206 & 0.9413 & 0.8732 & 0.9060\\
    \bottomrule

  \end{tabular}
  \caption{Retrieval performance of \texttt{REL} and \texttt{EXT}.\label{tbl:cls_rst}}
\end{table}

Next we discuss the main results comparing against the top two teams (rows 1--2)  in the 2019 track in Table~\ref{tbl:quant_results}.
Before we proceed, we want to highlight one crucial evaluation consideration that applies to any TREC track.
TREC  evaluates systems in the \textit{Cranfield} paradigm where pooled top documents from all participating teams are judged for relevance by human experts. Because we did not participate in the original TREC-PM 2019 task, our retrieved results are not part of the judged documents.  Hence, we may be at a slight disadvantage when comparing our results with those of teams that participated in 2019 TREC-PM. Nevertheless, we believe that at least the top few most relevant documents are typically commonly retrieved by all models. Hence we compare with both P@10 and R-Prec (P@all-relevant-doc-count) measures.

\begin{table}[bht]
  \small
  \centering
  \renewcommand{\arraystretch}{1.3}
  \setlength{\tabcolsep}{3pt}
  \begin{tabular}{@{}lcc@{}}
    \toprule
    Model & R-Prec & P@10 \\
    \midrule
    julie-mug~\cite{faessler2020julie} & 0.3572 & 0.6525 \\
    BITEM\_PM~\cite{caucheteur202designing} & 0.3166 & 0.6275 \\
    \midrule
   Baseline: Solr eDisMax & 0.2307 & 0.5200 \\
   Baseline + Solr MLT & 0.1773 & 0.2625 \\
   \midrule
    Baseline + \texttt{REL} & \textbf{0.3912} & 0.6750 \\
     Baseline + \texttt{ABS }& 0.2700 & 0.5625 \\
     Baseline + \texttt{REL}+\texttt{ABS} & 0.3627 & \textbf{0.6985} \\
    \bottomrule
  \end{tabular}
  \caption{Our scores and top entries in 2019 TREC-PM.\label{tbl:quant_results}}
\end{table}

Our baseline Solr query results are shown in row 3  with subsequent rows showing results from additional components. Solr \textit{eDisMax} is a document ranking function which is based on the BM25 \cite{jones2000probabilistic} probabilistic model. We also evaluate \textit{eDisMax} with Solr MLT (\texttt{MoreLikeThis}), in which a new query is generated by adding a few ``interesting'' terms (top TF/IDF terms) from the retrieved documents of the initial \textit{eDisMax}  query.  This traditional relevance feedback method (row 4) method has decreased the performance from the baseline and hence has not been used in our reranking methods.

All our models (rows 5--7) present stable baseline scores in P@10 and the combined method (\texttt{+REL+ABS}) tops the list with a 4\% improvement over the prior best model~\cite{faessler2020julie}.
Baseline with \texttt{REL} does the best in terms of R-Prec.
Both prior top teams rely heavily on query expansion through external knowledge bases to add synonyms, hypernyms, and hyponyms of terms found in the original query.

\subsection{Qualitative Analysis}


\begin{table*}[tbh]
  \centering
  \small
  \renewcommand{\arraystretch}{1.3}
  \begin{tabular}{@{}llp{7cm}@{}}
    \toprule
    Document & Facet signal & Summary \\
    \midrule
    \multirow{4}{6cm}{\textbf{Title:} Association between BRAF v600e mutation and the clinicopathological
      features of solitary papillary thyroid microcarcinoma. (PMID: 28454296)}
             & \texttt{[unused\_0]} & papillary intrahepatic cholangiocarcinoma \\
    {} & \texttt{[unused\_1]} & braf v600e \\
    {} & \texttt{[unused\_3]} & D018281 C535533 \\
    {} & \texttt{[unused\_4]} & papillary thyroid braf clinicopathological v600e \\
    \midrule
    \multirow{4}{6cm}{\textbf{Title:} Identification of differential and functionally active miRNAs in both anaplastic lymphoma kinase (ALK)+ and ALK- anaplastic large-cell lymphoma. (PMID: 20805506)}
             & \texttt{[unused\_0]} & lymphoma \\
    {} & \texttt{[unused\_1]} & anaplastic lymphoma alk cell bradykinin \\
    {} & \texttt{[unused\_3]} & D002471 D017728 D000077548\\
    {} & \texttt{[unused\_4]} & lymphoma alk receptor tyrosine kinase\\
    \bottomrule
  \end{tabular}
  \caption{Sample facet-conditioned document summarizations by \texttt{ABS}\label{tbl:sample_summaries}}
\end{table*}

Table~\ref{tbl:sample_summaries} presents sample pseudo-queries generated by \texttt{ABS}. The summaries of the first document show some novel words, \textit{intrahepatic} and \textit{cholangiocarcinoma}, that do not occur in the given document (we only show title for conciseness, but the abstract also does not contain those words). The model may have learned the close relationship between \textit{cholangiocarcinoma} and \textit{BRAF v600e}, the latter being part of the genetic facet of the actual query for which PMID: 28454296 turns out to be relevant. Also embedding proximity between \textit{intrahepatic} and \textit{cholangiocarcinoma} may have introduced both into the pseudo query, although they are not central to this document's theme. Still, this maybe important in retrieving documents that have an indirect (yet relevant) link to the query through the pseudo-query terms. This could be why, although \texttt{ABS} underperforms \texttt{REL}, it still complements it when combined (Table~\ref{tbl:quant_results}).
The table also shows that \texttt{ABS} can generate concepts in a domain-specific terminology. For example, the second document yields following MeSH entity codes, which are strongly related to the topics of the document: \textit{D002471} (Cell Transformation, Neoplastic), \textit{D017728} (Lymphoma, Large-Cell, Anaplastic), and \textit{D000077548} (Anaplastic Lymphoma Kinase).

For a qualitative exploration of what \texttt{EXT} and different facets of \texttt{ABS} capture, we refer the reader to Appendix~\ref{sec-appendix}.

\subsection{Machine Configuration and Runtime}
All  training and testing was done on a single Nvidia Titan X GPU in a desktop with 64GB RAM.
The corpus to be indexed had 30,429,310 biomedical citations (titles and abstracts of biomedical articles\footnote{Due to copyright issues with full-text, TREC-PM is only conducted on abstracts/titles of  articles available on PubMed.}).
We trained the three models for five epochs and the training time per epoch (80,319 query, doc pairs) is 69 mins for \texttt{REL}, 72 mins for \texttt{EXT},  and 303 mins for  \texttt{ABS}. Coming to test time,
per query, the Solr eDisMax query returns top 500 results in 20~ms. Generating pseudo-queries for 500 candidates via \texttt{EXT} and \texttt{ABS} takes 126 seconds and generating \texttt{REL} scores consumes 16 seconds.
So per query, it takes nearly 2.5 mins at test time to return a ranked list of documents. Although this does not facilitate real time retrieval as in commercial search engines, given the complexity of the queries, we believe this is at least near real time offering a convenient way to launch PM queries. Furthermore, this comes at an affordable configuration for many labs and clinics with a smaller carbon footprint.

\section{Conclusion}

In this paper, we  proposed an ensemble document reranking approach  for PM queries. It builds on pretrained BERT models to combine strategies from document relevance matching and extractive/abstractive text summarization to arrive at document rankings that are complementary in eventual evaluations. Our experiments also demonstrate that  entity embeddings   trained on an annotated domain specific corpus can help in   document retrieval settings. Both quantitative and qualitative analyses throw light on the strengths of our approach.

One scope for advances lies in improving the summarizer to generate better pseudo-queries  so that \texttt{ABS} starts to perform better on its own. At a high level, training data is very hard to generate in large amounts for IR tasks in biomedicine and this holds for the TREC-PM datasets too. To better train  \texttt{ABS}, it may be better to adapt other biomedical IR datasets. For example, the TREC clinical decision support (CDS) task that ran from 2014 to 2016 is related to the PM task~\cite{roberts2016state}. A future goal is to see if we can apply our neural transfer learning~\cite{rios2019neural} and domain adaptation~\cite{rios2018generalizing} efforts to repurpose the CDS datasets for the PM task.

Another straightforward idea is to reuse generated pseudo-query sentences in the eDisMax query by Solr, as a form of  pseudo relevance feedback. The $s_{\cos}$ expression in Section~\ref{sec-rerank} focuses on an asymmetric formulation that starts with a query term and looks for the best match in the pseudo-query. Considering a more symmetric formulation, where, we also begin with the pseudo-query terms and average both summands may provide a better estimate for reranking.
 Additionally, a thorough exploration of how external biomedical knowledge bases~\cite{wagner2020harmonized} can be incorporated in the neural IR framework for PM is also important~\cite{nguyen2017dsrim}.

\bibliographystyle{acl_natbib}
\bibliography{rt2001}{}

\begin{thebibliography}{35}
\expandafter\ifx\csname natexlab\endcsname\relax\def\natexlab#1{#1}\fi

\bibitem[{Bojanowski et~al.(2017)Bojanowski, Grave, Joulin, and
  Mikolov}]{bojanowski2017enriching}
Piotr Bojanowski, Edouard Grave, Armand Joulin, and Tomas Mikolov. 2017.
\newblock \href {https://doi.org/10.1162/tacl_a_00051} {Enriching word vectors
  with subword information}.
\newblock \emph{Transactions of the Association for Computational Linguistics},
  5:135--146.

\bibitem[{Caucheteur et~al.(2020)Caucheteur, Pasche, Gobeill, Mottaz, Mottin,
  and Ruch}]{caucheteur202designing}
Deborah Caucheteur, Emilie Pasche, Julien Gobeill, Anais Mottaz, Luc Mottin,
  and Patrick Ruch. 2020.
\newblock \href {https://trec.nist.gov/pubs/trec28/papers/BITEM_PM.PM.pdf}
  {Designing retrieval models to contrast precision-driven ad hoc search vs.
  recall-driven treatment extraction in precision medicine}.

\bibitem[{Cheng and Lapata(2016)}]{cheng2016neural}
Jianpeng Cheng and Mirella Lapata. 2016.
\newblock \href {https://doi.org/10.18653/v1/p16-1046} {Neural summarization by
  extracting sentences and words}.
\newblock In \emph{Proceedings of the 54th Annual Meeting of the Association
  for Computational Linguistics (Volume 1: Long Papers)}, pages 484--494.

\bibitem[{Collins and Varmus(2015)}]{collins2015new}
Francis~S Collins and Harold Varmus. 2015.
\newblock \href {https://doi.org/10.1056/NEJMp1500523} {A new initiative on
  precision medicine}.
\newblock \emph{New England journal of medicine}, 372(9):793--795.

\bibitem[{Conneau and Lample(2019)}]{conneau2019cross}
Alexis Conneau and Guillaume Lample. 2019.
\newblock Cross-lingual language model pretraining.
\newblock In \emph{Advances in Neural Information Processing Systems}, pages
  7057--7067.

\bibitem[{Cormack et~al.(2009)Cormack, Clarke, and
  Buettcher}]{cormack2009reciprocal}
Gordon~V Cormack, Charles~LA Clarke, and Stefan Buettcher. 2009.
\newblock \href {https://doi.org/10.1145/1571941.1572114} {Reciprocal rank
  fusion outperforms condorcet and individual rank learning methods}.
\newblock In \emph{Proceedings of the 32nd international ACM SIGIR conference
  on Research and development in information retrieval}, pages 758--759.

\bibitem[{Devlin et~al.(2019)Devlin, Chang, Lee, and
  Toutanova}]{devlin2019bert}
Jacob Devlin, Ming-Wei Chang, Kenton Lee, and Kristina Toutanova. 2019.
\newblock \href {https://doi.org/10.18653/v1/N19-1423} {{BERT}: Pre-training of
  deep bidirectional transformers for language understanding}.
\newblock In \emph{NAACL-HLT}, pages 4171--4186.

\bibitem[{Faessler et~al.(2020)Faessler, Oleynik, and Hahn}]{faessler2020julie}
Erik Faessler, Michel Oleynik, and Udo Hahn. 2020.
\newblock \href {https://trec.nist.gov/pubs/trec28/papers/julie-mug.PM.pdf}
  {{JULIE} lab \& {Med Uni Graz} @ {TREC} 2019 precision medicine track}.

\bibitem[{Grainger and Potter(2014)}]{grainger2014solr}
Trey Grainger and Timothy Potter. 2014.
\newblock \emph{Solr in action}.
\newblock Manning Publications Co.

\bibitem[{He et~al.(2016)He, Zhang, Ren, and Sun}]{he2016deep}
Kaiming He, Xiangyu Zhang, Shaoqing Ren, and Jian Sun. 2016.
\newblock \href {https://doi.org/10.1109/cvpr.2016.90} {Deep residual learning
  for image recognition}.
\newblock In \emph{Proceedings of the IEEE conference on computer vision and
  pattern recognition}, pages 770--778.

\bibitem[{Jones et~al.(2000)Jones, Walker, and
  Robertson}]{jones2000probabilistic}
K~Sparck Jones, Steve Walker, and Stephen~E. Robertson. 2000.
\newblock A probabilistic model of information retrieval: development and
  comparative experiments: Part 2.
\newblock \emph{Information processing \& management}, 36(6):809--840.

\bibitem[{Klein et~al.(2017)Klein, Kim, Deng, Senellart, and
  Rush}]{klein2017opennmt}
Guillaume Klein, Yoon Kim, Yuntian Deng, Jean Senellart, and Alexander~M Rush.
  2017.
\newblock \href {https://doi.org/10.18653/v1/p17-4012} {Opennmt: Open-source
  toolkit for neural machine translation}.
\newblock In \emph{Proceedings of ACL 2017, System Demonstrations}, pages
  67--72.

\bibitem[{Lin(2004)}]{lin-2004-rouge}
Chin-Yew Lin. 2004.
\newblock \href {https://www.aclweb.org/anthology/W04-1013} {{ROUGE}: A package
  for automatic evaluation of summaries}.
\newblock In \emph{Text Summarization Branches Out}, pages 74--81, Barcelona,
  Spain. Association for Computational Linguistics.

\bibitem[{Liu and Lapata(2019)}]{liu2019text}
Yang Liu and Mirella Lapata. 2019.
\newblock \href {https://doi.org/10.18653/v1/d19-1387} {Text summarization with
  pretrained encoders}.
\newblock In \emph{Proceedings of the 2019 Conference on Empirical Methods in
  Natural Language Processing and the 9th International Joint Conference on
  Natural Language Processing (EMNLP-IJCNLP)}, pages 3721--3731.

\bibitem[{Mueller and Thyagarajan(2016)}]{mueller2016siamese}
Jonas Mueller and Aditya Thyagarajan. 2016.
\newblock Siamese recurrent architectures for learning sentence similarity.
\newblock In \emph{Proceedings of the Thirtieth AAAI Conference on Artificial
  Intelligence}, pages 2786--2792.

\bibitem[{Nallapati et~al.(2017)Nallapati, Zhai, and
  Zhou}]{nallapati2017summarunner}
Ramesh Nallapati, Feifei Zhai, and Bowen Zhou. 2017.
\newblock Summarunner: A recurrent neural network based sequence model for
  extractive summarization of documents.
\newblock In \emph{Thirty-First AAAI Conference on Artificial Intelligence}.

\bibitem[{Nallapati et~al.(2016)Nallapati, Zhou, dos Santos, Gul{\c{c}}ehre,
  and Xiang}]{nallapati2016abstractive}
Ramesh Nallapati, Bowen Zhou, Cicero dos Santos, {\c{C}}a{\u{g}}lar
  Gul{\c{c}}ehre, and Bing Xiang. 2016.
\newblock \href {https://doi.org/10.18653/v1/k16-1028} {Abstractive text
  summarization using sequence-to-sequence rnns and beyond}.
\newblock In \emph{Proceedings of The 20th SIGNLL Conference on Computational
  Natural Language Learning}, pages 280--290.

\bibitem[{Narayan et~al.(2018)Narayan, Cohen, and Lapata}]{narayan2018ranking}
Shashi Narayan, Shay~B Cohen, and Mirella Lapata. 2018.
\newblock \href {https://doi.org/10.18653/v1/n18-1158} {Ranking sentences for
  extractive summarization with reinforcement learning}.
\newblock In \emph{NAACL-HLT}, pages 1747--1759.

\bibitem[{Nguyen et~al.(2017)Nguyen, Soulier, Tamine, and
  Bricon-Souf}]{nguyen2017dsrim}
Gia-Hung Nguyen, Laure Soulier, Lynda Tamine, and Nathalie Bricon-Souf. 2017.
\newblock \href {https://doi.org/10.1145/3209978.3210081} {Dsrim: A deep neural
  information retrieval model enhanced by a knowledge resource driven
  representation of documents}.
\newblock In \emph{Proceedings of the ACM SIGIR International Conference on
  Theory of Information Retrieval}, pages 19--26.

\bibitem[{Nogueira and Cho(2017)}]{nogueira2017task}
Rodrigo Nogueira and Kyunghyun Cho. 2017.
\newblock \href {https://doi.org/10.18653/v1/d17-1061} {Task-oriented query
  reformulation with reinforcement learning}.
\newblock In \emph{Proceedings of the 2017 Conference on Empirical Methods in
  Natural Language Processing}, pages 574--583.

\bibitem[{Noh and Kavuluru(2018)}]{noh2018document}
Jiho Noh and Ramakanth Kavuluru. 2018.
\newblock \href {https://doi.org/10.1109/icmla.2018.00036} {Document retrieval
  for biomedical question answering with neural sentence matching}.
\newblock In \emph{2018 17th IEEE International Conference on Machine Learning
  and Applications (ICMLA)}, pages 194--201. IEEE.

\bibitem[{Paulus et~al.(2018)Paulus, Xiong, and Socher}]{paulus2017deep}
Romain Paulus, Caiming Xiong, and Richard Socher. 2018.
\newblock A deep reinforced model for abstractive summarization.
\newblock \emph{International Conference on Learning Representations}.

\bibitem[{Prasad and Kan(2019)}]{prasad2019glocal}
Animesh Prasad and Min-Yen Kan. 2019.
\newblock \href {https://doi.org/10.18653/v1/N19-1182} {Glocal: Incorporating
  global information in local convolution for keyphrase extraction}.
\newblock In \emph{NAACL-HLT}, pages 1837--1846.

\bibitem[{Rios and Kavuluru(2019)}]{rios2019neural}
Anthony Rios and Ramakanth Kavuluru. 2019.
\newblock \href {https://doi.org/10.1016/j.artmed.2019.04.002} {Neural transfer
  learning for assigning diagnosis codes to emrs}.
\newblock \emph{Artificial Intelligence in Medicine}, 96:116--122.

\bibitem[{Rios et~al.(2018)Rios, Kavuluru, and Lu}]{rios2018generalizing}
Anthony Rios, Ramakanth Kavuluru, and Zhiyong Lu. 2018.
\newblock \href {https://doi.org/10.1093/bioinformatics/bty190} {Generalizing
  biomedical relation classification with neural adversarial domain
  adaptation}.
\newblock \emph{Bioinformatics}, 34(17):2973--2981.

\bibitem[{Roberts et~al.(2020)Roberts, Demner-Fushman, Voorhees, Hersh,
  Bedrick, Lazar, Pant, and Meric-Bernstam}]{roberts2020overview}
Kirk Roberts, Dina Demner-Fushman, Ellen~M. Voorhees, William~R. Hersh, Steven
  Bedrick, Alexander~J. Lazar, Shubham Pant, and Funda Meric-Bernstam. 2020.
\newblock \href {https://trec.nist.gov/pubs/trec28/papers/OVERVIEW.PM.pdf}
  {Overview of the {TREC} 2019 precision medicine track}.

\bibitem[{Roberts et~al.(2016)Roberts, Simpson, Demner-Fushman, Voorhees, and
  Hersh}]{roberts2016state}
Kirk Roberts, Matthew Simpson, Dina Demner-Fushman, Ellen Voorhees, and William
  Hersh. 2016.
\newblock \href {https://doi.org/10.1007/s10791-015-9259-x} {State-of-the-art
  in biomedical literature retrieval for clinical cases: a survey of the trec
  2014 cds track}.
\newblock \emph{Information Retrieval Journal}, 19(1):113--148.

\bibitem[{Rush et~al.(2015)Rush, Chopra, and Weston}]{rush2015neural}
Alexander~M Rush, Sumit Chopra, and Jason Weston. 2015.
\newblock \href {https://doi.org/10.18653/v1/d15-1044} {A neural attention
  model for abstractive sentence summarization}.
\newblock In \emph{EMNLP}, pages 379--389.

\bibitem[{Schuster and Nakajima(2012)}]{schuster2012japanese}
Mike Schuster and Kaisuke Nakajima. 2012.
\newblock Japanese and korean voice search.
\newblock In \emph{2012 IEEE International Conference on Acoustics, Speech and
  Signal Processing (ICASSP)}, pages 5149--5152. IEEE.

\bibitem[{See et~al.(2017)See, Liu, and Manning}]{see2017get}
Abigail See, Peter~J Liu, and Christopher~D Manning. 2017.
\newblock \href {https://doi.org/10.18653/v1/p17-1099} {Get to the point:
  Summarization with pointer-generator networks}.
\newblock In \emph{Proceedings of the 55th Annual Meeting of the Association
  for Computational Linguistics}, pages 1073--1083.

\bibitem[{Sun et~al.(2019)Sun, Tang, Du, Deng, and
  Nie}]{sun2019divgraphpointer}
Zhiqing Sun, Jian Tang, Pan Du, Zhi-Hong Deng, and Jian-Yun Nie. 2019.
\newblock \href {https://doi.org/10.1145/3331184.3331219} {Divgraphpointer: A
  graph pointer network for extracting diverse keyphrases}.
\newblock In \emph{Proceedings of the 42nd International ACM SIGIR Conference
  on Research and Development in Information Retrieval}, pages 755--764.

\bibitem[{Vaswani et~al.(2017)Vaswani, Shazeer, Parmar, Uszkoreit, Jones,
  Gomez, Kaiser, and Polosukhin}]{vaswani2017attention}
Ashish Vaswani, Noam Shazeer, Niki Parmar, Jakob Uszkoreit, Llion Jones,
  Aidan~N Gomez, {\L}ukasz Kaiser, and Illia Polosukhin. 2017.
\newblock Attention is all you need.
\newblock In \emph{Advances in neural information processing systems}, pages
  5998--6008.

\bibitem[{Wagner et~al.(2020)Wagner, Walsh, Mayfield, Tamborero, Sonkin,
  Krysiak, Deu-Pons, Duren, Gao, McMurry et~al.}]{wagner2020harmonized}
Alex~H Wagner, Brian Walsh, Georgia Mayfield, David Tamborero, Dmitriy Sonkin,
  Kilannin Krysiak, Jordi Deu-Pons, Ryan~P Duren, Jianjiong Gao, Julie McMurry,
  et~al. 2020.
\newblock \href {https://doi.org/10.1101/366856} {A harmonized
  meta-knowledgebase of clinical interpretations of somatic genomic variants in
  cancer}.
\newblock \emph{Nature genetics}, 52(4):448--457.

\bibitem[{Wolf et~al.(2019)Wolf, Debut, Sanh, Chaumond, Delangue, Moi, Cistac,
  Rault, Louf, Funtowicz et~al.}]{wolf2019transformers}
Thomas Wolf, Lysandre Debut, Victor Sanh, Julien Chaumond, Clement Delangue,
  Anthony Moi, Pierric Cistac, Tim Rault, R{\'e}mi Louf, Morgan Funtowicz,
  et~al. 2019.
\newblock Transformers: State-of-the-art natural language processing.
\newblock \emph{arXiv preprint arXiv:1910.03771}.

\bibitem[{Wu et~al.(2016)Wu, Schuster, Chen, Le, Norouzi, Macherey, Krikun,
  Cao, Gao, Macherey et~al.}]{wu2016google}
Yonghui Wu, Mike Schuster, Zhifeng Chen, Quoc~V Le, Mohammad Norouzi, Wolfgang
  Macherey, Maxim Krikun, Yuan Cao, Qin Gao, Klaus Macherey, et~al. 2016.
\newblock Google's neural machine translation system: Bridging the gap between
  human and machine translation.
\newblock \emph{arXiv preprint arXiv:1609.08144}.

\end{thebibliography}

\appendix
\section{Attention Heatmaps by Facet Signals}
\label{sec-appendix}

\begin{figure*}[h]
   \centering
   \includegraphics[width=\textwidth]{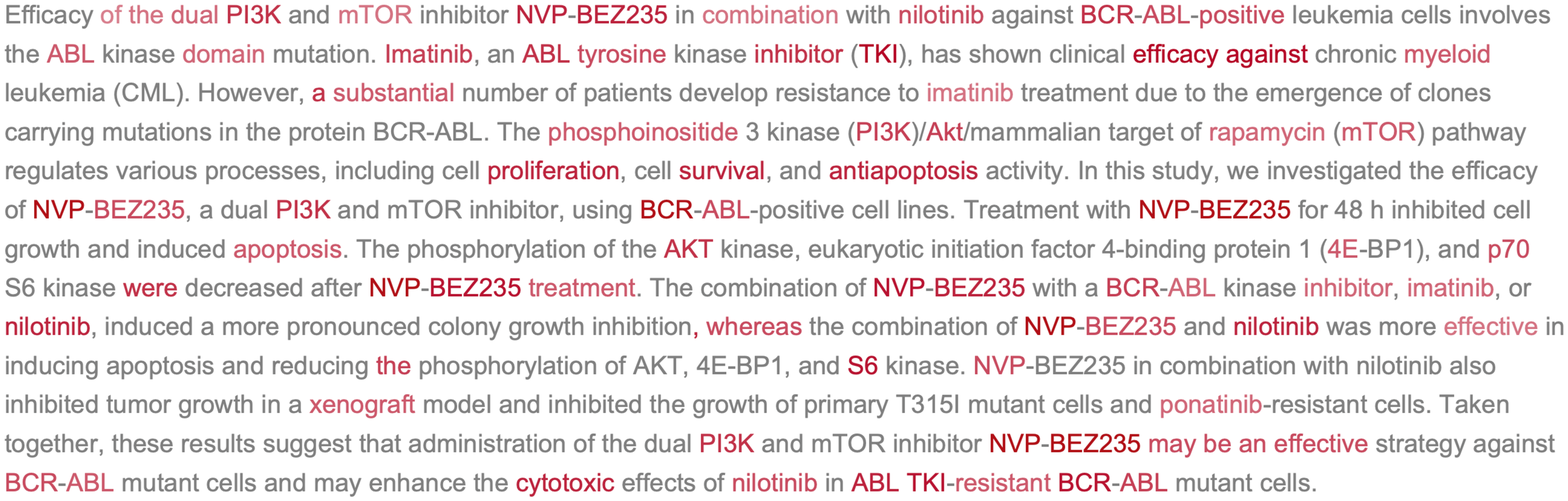}
   \caption{Heatmap of classification scores by \texttt{EXT}. Darker red indicates relatively higher probability of the token being relevant to the theme of the TREC-PM datasets.\label{fig:ext_heatmap}}
\end{figure*}

\begin{figure*}[h]
  \centering
  \begin{subfigure}[b]{\textwidth}
  \includegraphics[width=\textwidth]{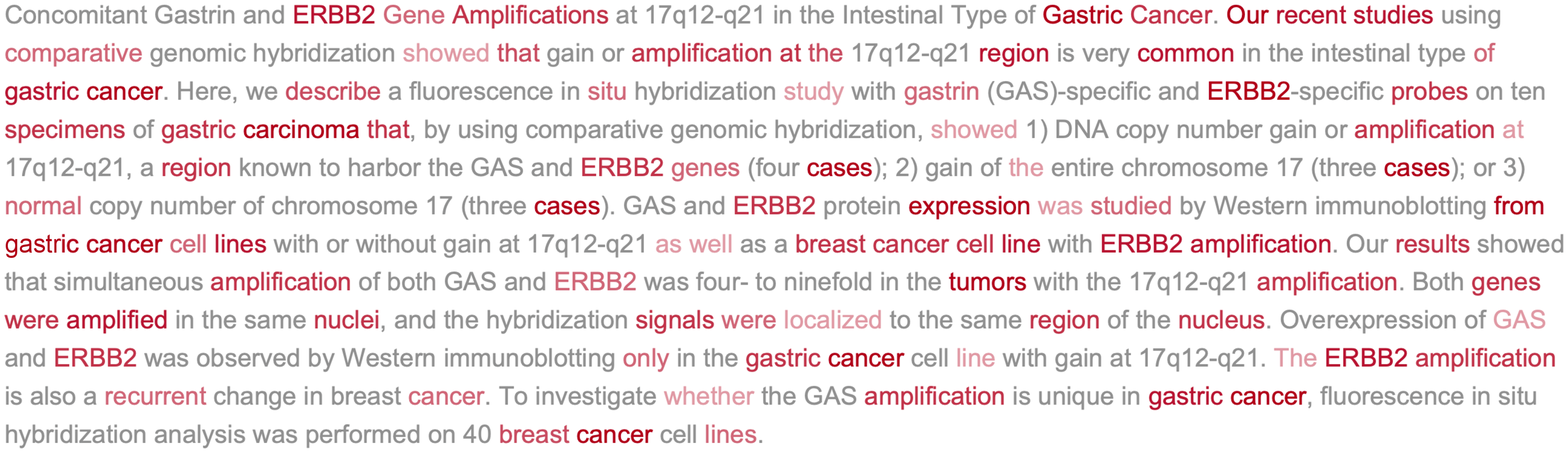}
  \caption{Attention heatmap produced by \texttt{[unused0]} signal (topic of disease)}
  \end{subfigure}

  \begin{subfigure}[b]{\textwidth}
  \includegraphics[width=\textwidth]{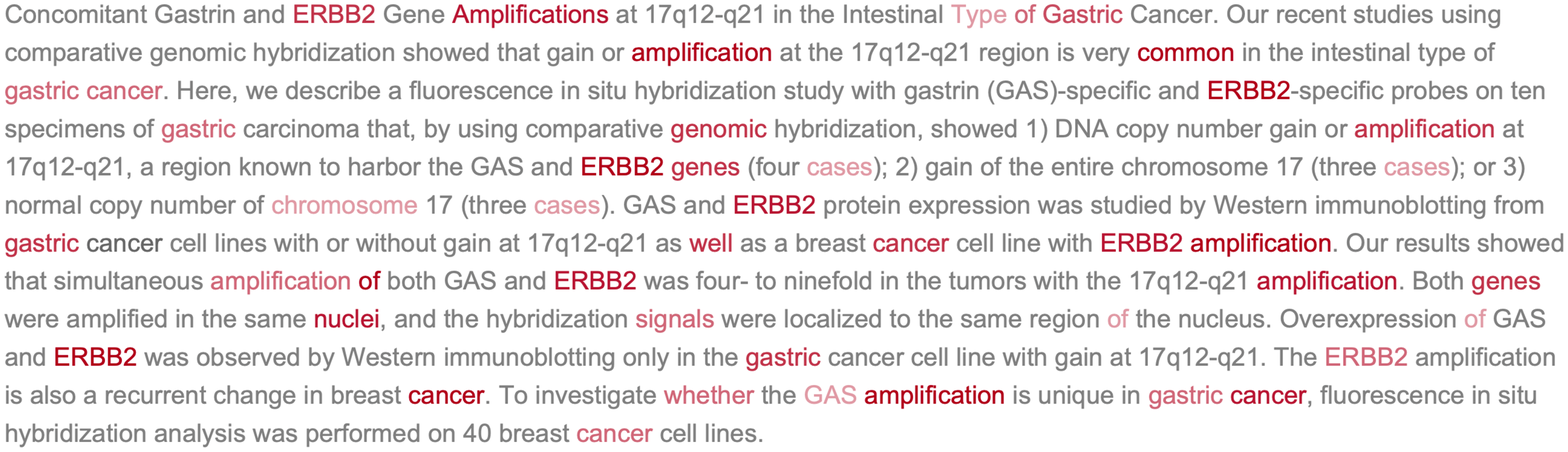}
  \caption{Attention heatmap produced by \texttt{[unused1]} signal (topic of generic variants and gene regulations)}
  \end{subfigure}

  \caption{Comparison between the attention heatmaps on a sample document conditioned by field signals in \texttt{ABS} model.~\label{fig:abs_heatmap}}
\end{figure*}

Figure~\ref{fig:ext_heatmap} depicts words highlighted by \texttt{EXT}. Evidently, we see terms  related to the regulations of gene expressions, proteins, or disease names featuring more prominently. Figure~\ref{fig:abs_heatmap} shows how \texttt{ABS} reads the source document differently depending on which facet signal it starts with, in the process of query generation; compared to \texttt{[unused0]} (disease facet), the attention heat map by \texttt{[unused1]} (genetic facet) focuses more on the words related to gene regulations.

\end{document}